\documentclass[runningheads]{llncs}

\usepackage{url}
\usepackage{bbold}
\usepackage{multirow}
\usepackage{graphicx, nicefrac}
\usepackage{booktabs}
\usepackage{amsfonts}
\usepackage{amsmath}
\usepackage{enumitem}
\usepackage{comment}

\usepackage[sort]{cite}
\usepackage[hidelinks]{hyperref}
\usepackage[misc]{ifsym}
\usepackage{xcolor}
\usepackage[linesnumbered,vlined,ruled]{algorithm2e}
\usepackage{orcidlink}

\usepackage{tikz}
\usetikzlibrary{decorations.pathreplacing,calc}
\newcommand{\tikzmark}[1]{\tikz[overlay,remember picture] \node (#1) {};}

\newcommand{\bridget}{\textsc{Bridget}}

\definecolor{darkgreen}{rgb}{0.0, 0.5, 0.0}
\definecolor{darkblue}{rgb}{0.0, 0.2, 0.6}
\definecolor{darkpastelred}{rgb}{0.76, 0.23, 0.13}
\definecolor{carrotorange}{rgb}{0.93, 0.57, 0.13}

\begin{document}

\setlength{\abovedisplayskip}{3pt}
\setlength{\belowdisplayskip}{3pt}

\title{Bridging the Gap\\in Hybrid Decision-Making Systems}
\titlerunning{Bridging the Gap in Hybrid Decision-Making Systems}

\author{Federico Mazzoni\inst{1}\orcidlink{0009-0004-6961-7495},
    Roberto Pellungrini\inst{2}\orcidlink{0000-0003-3268-9271},
\and
    Riccardo Guidotti\inst{1}\orcidlink{0000-0002-2827-7613} {\footnotesize{\Letter}}
    %Alessio Malizia\inst{1} \orcidlink{0000-0002-2601-7009} \and
    %Andrea Beretta\inst{2} \orcidlink{0000-0001-8531-9325} \\ \and
    %\and 
    %Anna Monreale\inst{1} \orcidlink{0000-0001-8541-0284}
}

\authorrunning{F. Mazzoni et al.}

%\tocauthor{Federico Mazzoni, Riccardo Guidotti, Alessio Malizia}
%\toctitle{A \frank{} System for Co-Evolutionary Hybrid Decision-Making}

\institute{
    $^1$University of Pisa, Italy, \email{federico.mazzoni@phd.unipi.it},
    \email{riccardo.guidotti@unipi.it},
    %\\ 
    %\email{\{name.surname\}@phd.unipi.it},
    %\email{\{name.surname\}@unipi.it},\\
    %$^2$Molde University College, Norway\\ 
    %$^2$ISTI-CNR Pisa, Italy, 
    %\email{\{name.surname\}@isti.cnr.it}, 
    %\\
    $^2$Scuola Normale Superiore, Pisa, Italy, \email{roberto.pellungrini@sns.it}
    %\email{\{name.surname\}@sns.it}
}

%\tocauthor{Federico Mazzoni, Riccardo Guidotti, Alessio Malizia}
%\toctitle{A \frank{} System for Co-Evolutionary Hybrid Decision-Making}

\authorrunning{F. Mazzoni et al.}

\maketitle

\begin{abstract}
We introduce \bridget{}, a novel human-in-the-loop system for hybrid decision-making, aiding the user to label records from an un-labeled dataset, attempting to ``bridge the gap'' between the two most popular Hybrid Decision-Making paradigms: those featuring the human in a leading position, and the other with a machine making most of the decisions.
\bridget{} understands when either a machine or a human user should be in charge, dynamically switching between two statuses. 
In the different statuses, \bridget{} still fosters the human-AI interaction, either having a machine learning model assuming \textit{skeptical} stances towards the user and offering them suggestions, or towards itself and calling the user back.
We believe our proposal lays the groundwork for future synergistic systems involving a human and a machine decision-makers.

\end{abstract}

\section{Introduction}
Automated decision-making processes based on Machine Learning (ML) are still not widely adopted for high-stakes decisions such as medical diagnoses or court decisions~\cite{wang2020augmented}.
In these fields, humans are aided but not replaced by Artificial Intelligence (AI), resulting in Hybrid Decision-Makers (HDM)~\cite{mosier2018human}.

Ash the literature keeps growing, the term ``Hybrid Decision-Makers'' has been used as an umbrella word for various different kinds of algorithms, often with a different focus, and a proper consensus has not been reached yet. Punzi et al.~\cite{punzi2024ai} notes two major HDM paradigms: \textit{Learning-to-Abstain} where under certain conditions an ML model refuses to make a decision, and \textit{Learning Together}, where the human can interact with the training process of the ML model. Two of the most representative approaches of the two paradigms are, respectively, \textit{Learning-to-Defer} (LtD)~\cite{joshi2021learning} systems, where the machine plays the primary role, deferring decisions on records with a high degree of uncertainty to an external human supervisor,
%In~\cite{wang2020augmented}, a rule-based AI model with inferred rules suggests replacing some user's decisions to maximize fairness, whereas in~\cite{jarrett2022online}, the model mediates between a user and their supervisor if it is not confident in the user's decisions.
and \textit{Skeptical Learning} (SL), where an ML model learns ``in parallel'' to the decisions taken by a human and queries them if it is ``skeptical'' of the human decision~\cite{zhang2019personal,zeni2019fixing}.
The two have vastly different scopes.
LtD assumes a cost to query the human user and aims to minimize that, leaving most of the decisions to the machine, while SL assumes the human user is always in control but needs to be helped by a machine model to stay consistent over time. 
Whereas SL posits a human expert who always has the final say but who can albeit get confused and thus in need of the machine's help, LtD assumes that some decisions are better suitable either to the user or the machine. 
As such, the model is trained not only to classify instances but to defer them~\cite{punzi2024ai}.
Another key difference is the training phase. 
SL training is continuous, i.e., the model is always learning something from the user's final decision, acting as ground truth, whereas LtD employs a stationary dataset, thus resulting in a stationary model, with different ground truths for the classification and the deferral policy.
Therefore, HDM systems can also be classified either following the role of the primary decision-maker or their training process. %, effectively leading to alternate taxonomies from the one proposed in~\cite{punzi2024ai}.
Although Explanatory Interactive ML systems have been proposed~\cite{teso2019explanatory}, HDMs mostly focus on providing decisions rather than explanations. With that said, \textsc{CINCER} can provide explanations as contrastive and influential counterexamples~\cite{bontempelli2020learning}, whereas
\textsc{FRANK} can show the model logic and provide them with real and synthetic examples and counterexamples~\cite{Mazzoni2024frank}. Both are based on SL. %evolving over time as the model learns from the user 
%Also, neither of them typically accounts for explainability aspects, mainly focusing on providing the user with the decision rather than on the reasons for the decision~\cite{guidotti2018survey,Mazzoni2024frank}.

%SL aims to help the user remain consistent with their past decisions, still giving them veto power against the model's suggestions.
%%SL has been extensively applied to personal context recognition~\cite{bontempelli2020learning, zhang2019personal} and image classifications~\cite{teso2021interactive}. In~\cite{teso2021interactive}, \textsc{SL} suggestions are also supported by \textit{contrastive explanations}. 
Our proposal aims to ``fill the gap'' between human-driven HDMs, such as SL, and machine-led ones, such as LtD, effectively creating a ``bridge'', hence the name \bridget, between the two paradigms, and an interpretable system able to adapt to different scenarios.
%and machine employs and extends traditional SL, by taking into account simultaneously fairness aspects, explainable suggestions, and the involvement of the user's supervisor. 
Following~\cite{bontempelli2020learning,Mazzoni2024frank}, \bridget{} employs an \textit{Incremental Learning} (IL, or Continual Learning) model. IL is a ML paradigm where the model is continuously trained on small data batches, potentially only one data point, instead of the entirety of the training set~\cite{de2021continual,wang2023comprehensive}. Moreover, \bridget{} shares a focus on explanation with other interactive ML methods~\cite{banerjee2024learning,Mazzoni2024frank, teso2021interactive,teso2019explanatory}.

\section{Setting the Stage}
\label{sec:background}
%We keep the paper self-contained by reporting in the following 
In the following we report a brief overview of concepts necessary to understand our proposal. 
We indicate with $H$ and $M$ the Human user and the Machine of the system, and with $X, Y$ a dataset where $X\in\mathcal{X}$
%$X = \{x_1, \dots, x_n\} \in \mathcal{X}^{(m)}$
is a set of $n$ records in feature space $\mathcal{X}$, %i.e., $x_i = \{(a_1, v_1), \dots, (a_m, v_m)\}$, where $a_i$ is the attribute name and $v_i$ is the corresponding value, and $\mathcal{X}^{(m)}$ is the feature space consisting of $m$ input features,
while $Y \in \mathcal{Y}$
%$Y = \{y_1, \dots, y_n\} \in \mathcal{Y}$
is the set of the target variable in the target space $\mathcal{Y}$.
%With $A = \{a_1, \dots, a_m\}$ we indicate the set of feature names, and for an instance $x \in X$, we write $x[a_k]$ to refer to the value $v_k$ of attribute $a_k$.
For classification problems, $y_i \in \{1, \dots, l\} = L$, where $L$ is the set of different class labels and $l$, is the number of the classes. %, while when dealing with regression problems, $y_i \in \mathbb{R}$. 
%Without losing in generality, we consider $l=2$, i.e., binary classification problems.
We indicate a trained decision-making model with a function $f:\mathcal{X} \rightarrow \mathcal{Y}$ that maps data instances $x$ from the feature space $\mathcal{X}$ to the target space $\mathcal{Y}$. We represent the user decision process as an analogous function $h:\mathcal{X} \rightarrow \mathcal{Y}$.
We write $f(x_i) = \tilde{y_i}$ to denote the decision $\tilde{y_i}$ taken by $f$, $h(x_i) = \hat{y_i}$ to denote the decision $\hat{y_i}$  taken by $h$.
%and $f(X) = Y$ as a shorthand for $\{f(x_i) \ \vert \ x_i \in X\} = Y$. Here, $f$ is the classifier of the machine $M$ and $h$ represents the decision process of $H$.

\smallskip
\textbf{Skeptical Learning.} 
Given a ML model $f$ and a dataset $X$, the user is tasked to assign a label $y_i$ to each record $x_i \in X$. 
In SL, the user assigns the label $\hat{y}_i$ and, independently from them, $f$ assigns the label $\tilde{y}_i$. 
The ML model $f$ can be pre-trained on a small training set. 
If $\hat{y}_i \neq \tilde{y}_i$ and $f$ is \textit{skeptical} (see below), the user is asked if they want to accept $\tilde{y}_i$ as $y_i$. 
If they do, $y_i$ takes the value $\tilde{y}_i$. 
If the user refuses, if $\hat{y}_i = \tilde{y}_i$ or if the model is not skeptical, $y_i$ is assigned $\hat{y}_i$.
%The ML model 
$f$ is then incrementally trained on $x_i$ and $y_i$.

The model's \textit{skepticality} is related to the model's \textit{epistemic uncertainty}~\cite{teso2021interactive}, which is independent of the notion of prediction probability towards a certain decision.
%\textit{confidence} score towards a certain decision, i.e., the prediction probability.
%\footnote{\scriptsize Note that there's a general lack of normativity w.r.t. these terms; e.g.,~\cite{zeni2019fixing} uses the term \textit{confidence} to refer to the epistemic uncertainty.}. 
As the model is exposed to enough data, its epistemic uncertainty, i.e., its \textit{ignorance}, should be minimized, assuming consistency in the labels~\cite{hullermeier2021aleatoric}.
%Epistemic uncertainty is the model's \textit{ignorance}, and given enough data, it should be minimized~\cite{hullermeier2021aleatoric}.
Since few models allow access to epistemic uncertainty~\cite{hullermeier2021aleatoric,bontempelli2020learning}, it has been approximated by SL %implementations 
with the \textit{Empirical Accuracy} (EA) of past predictions, both of the user and the model, i.e., the ratio between the number of times a label has been proposed by the user/predicted by the model, and the times it has been accepted as $y_i$~\cite{zeni2019fixing}.
Let $X_p \subseteq X$ be the set of past-seen instances and $Y_p \subseteq Y$ the respective ground truths. Let $\delta_{f(x),y}$ be the Kronecker Delta measuring accurate prediction of $f(x)$ w.r.t ground truth $y$. We calculate Skepticality as:
%
%\begin{equation}
%    \label{eq:skept}
% $$
%     \mathit{skp}(x_i, \tilde{y}_i, \hat{y}_i, Y, \tilde{Y}, \hat{Y}) = \mathit{c}(f, x_i, \tilde{y}_i) \cdot \mathit{ea}(\tilde{y}_i, Y, \tilde{Y}) - \mathit{c}(f, x_i, \hat{y}_i) \cdot \mathit{ea}(\hat{y}_i, Y, \hat{Y})
% $$
$$
    \mathit{skp}(x_i, \tilde{y}_i, \hat{y}_i,X_p, Y_p) = \mathit{c}_f(x_i, \tilde{y}_i) \cdot \mathit{ea}_f(\tilde{y_i}, X_p, Y_p) - \mathit{c}_f(x_i, \hat{y}_i) \cdot \mathit{ea}_h(\hat{y_i}, X_p, Y_p)
$$
%\end{equation}
%
where $\mathit{c}_f(x_i, \tilde{y}_i)$ and $\mathit{c}_f(x_i, \hat{y}_i)$ are the model prediction probabilities towards $\tilde{y}_i$ and $\hat{y}_i$, and $\mathit{ea}_f(\tilde{y}_i, X_p, Y_p) = \frac{1}{|X_p|} \sum_{\{x_p \in X_p | f(x_p) =\tilde{y}_i\}}\delta_{f(x_p),y_p}$ is the empirical accuracy of the model $f$ toward label $\tilde{y}_i$, and $\mathit{ea}_h(\hat{y}_i, X_p, Y_p)$ = $\frac{1}{|X_p|}$ $\sum_{\{x_p \in X_p | h(x_p)=\hat{y}_i\}}$ $\delta_{h(x_p),y_p}$ is the empirical accuracy of the user function $h$ toward label $\hat{y}_i$.
%The EA is computed as the cardinality of the intersection between the subset of all their past decisions with label either $\hat{y}_i$ or $\tilde{y}_i$ and the corresponding subset in $Y$, i.e., the final decision, over the subset of all their past decisions with either $\hat{y}_i$ or $\tilde{y}_i$. 
Thus, each possible label $l \in L$ has two EA values -- following the user's and the model's track record (e.g., in binary classification with $|L| = 2$, we have %a total of 
4 EA values, 2 for $M$ and 2 for $H$)~\cite{zeni2019fixing}. 
%In~\cite{zeni2019fixing} $H$'s EA values are initialized with $1$, and $M$'s with $0$ preventing in this way the model of being skeptical of earlier decisions.

\begin{comment}
    
\textbf{Explanatory Interactive ML.} Interactive explanatory systems are a close field to HDMs~\cite{teso2019explanatory}. 
\textsc{CINCER} extends SL by offering explanations as contrastive and influential counter-examples~\cite{bontempelli2020learning}.
\textsc{FRANK} can show the model logic and provide them with real and synthetic examples and counterexamples, %evolving over time as the model learns from the user
~\cite{Mazzoni2024frank}.
\end{comment}

%Learning-to-Guide provides the user with understandable and informative hits in natural language, leading them towards the machine's prediction without overtly revealing it\cite{banerjee2024learning}.

\begin{figure}[t]
    \centering
    \includegraphics[width=0.6\textwidth]{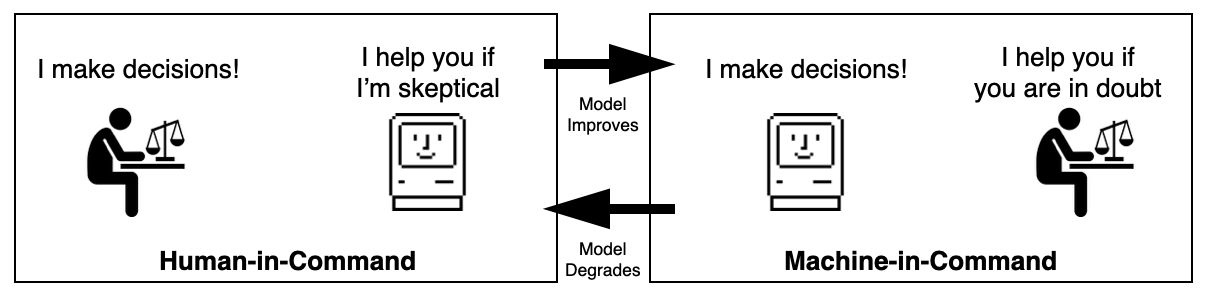} 
    \caption{\bridget{} looping between its potential phases.}
    \label{fig:bridget}
\end{figure}

\SetKwComment{Comment}{/* }{ */}

\begin{algorithm}[t]
\caption{\bridget{}}\label{alg:algoFrank}
    \SetKwInOut{Input}{Input}
    \SetKwInOut{Output}{Output}
    \Input{$X$ - records,
    $\alpha$ - skeptical threshold,
    $\beta$ - belief threshold
    }
    
    $X_p, Y_p, \tilde{Y}, \hat{Y}, f, k, p, phase \gets \mathit{initialize}$;
    %$FEA_M, FEA_H, p, k, f \gets \mathit{initialize}$;
    \tcp*[f]{\texttt{\scriptsize sets initialization}}\\
    $x_i \gets \mathit{receive\_record}(X)$;\tcp*[f]{\texttt{\scriptsize receive a new un-label record}}\\
    \uIf(\tcp*[f]{\texttt{\scriptsize if Human-in-Command}}){$\mathit{phase} = \mathit{HiC}$}{
        
        $\hat{y}_i \gets \mathit{h(x_i)}$; $\tilde{y}_i \gets \mathit{f(x_i)}$; \tcp*[f]{\texttt{\scriptsize get user decision and model prediction}}\\
        \uIf(\tcp*[f]{\texttt{\scriptsize if clash \& skepticism }}){$ \hat{y}_i \neq \tilde{y}_i \wedge \mathit{skp}(x_i, \tilde{y}_i, \hat{y}_i,X_p, Y_p) > \alpha$ } {

            \If(\tcp*[f]{\texttt{\scriptsize if an explanation for $\tilde{y}_i$ is desired}}){$\mathit{is\_expl\_desired(x_i, \tilde{y}_i)}$}{
                $e_i \gets \mathit{get\_and\_show\_expl}(x_i, \tilde{y}_i, f, X_p)$;\tcp*[f]{\texttt{\scriptsize return explanation $e_i$}}\\
            }

            \lIf(\tcp*[f]{\texttt{\scriptsize $\tilde{y}_i$ is accepted}}){$\mathit{accept\_label\_change}(x_i, \tilde{y}_i)$}{
                    $y_i \gets \tilde{y}_i$}
                \lElse(\tcp*[f]{\texttt{\scriptsize $\tilde{y}_i$ is refused}}){$y_i \gets \hat{y}_i$}
        }\tikzmark{skp_bottom}
        \lElse(\tcp*[f]{\texttt{\scriptsize otherwise $\hat{y}_i$ is accepted}}){$y_i \gets \hat{y}_i$}

    $\tilde{Y}, \hat{Y}, f, p \gets \mathit{update}$;
    \tcp*[f]{\texttt{\scriptsize model and parameter updates}}\\

    \lIf(\tcp*[f]{\texttt{\scriptsize phase change}}){$\mathit{check}(h,f, p)$}{
                    $phase = MiC, p = 0$}

}

    \Else(\tcp*[f]{\texttt{\scriptsize if Machine-in-Command}}){
    
    $\tilde{y}_i \gets \mathit{f(x_i)}$; $b \gets \mathit{compute}   (f, x_i, \tilde{y}_i, {Y}, \tilde{Y}$);
    \tcp*[f]{\texttt{\scriptsize $\tilde{y_i}$ and $M$'s belief towards it}}\\

    %%%%

    \uIf(\tcp*[f]{\texttt{\scriptsize if $b$ is low, user is called back}}){$\mathit{b < \beta}$}{
    $u_i \gets \mathit{get\_and\_show\_unr}(x_i, \tilde{y}_i, f, X_p)$;\tcp*[f]{\texttt{\scriptsize show explanation $u_i$}}\\
    $\hat{y}_i \gets \mathit{h(x_i)}$; $y_i \gets \hat{y}_i$;   
    \tcp*[f]{\texttt{\scriptsize $\hat{y}_i$ accepted}}\\
}

    \lElseIf(\tcp*[f]{\texttt{\scriptsize probablistic check, $\tilde{y}_i$ accepted}}){$\mathit{random\_check(b)}$}{
                    $y_i \gets \tilde{y}_i$} 
                    %%%%%%
    \Else(
    \tcp*[f]
    {\texttt{\scriptsize if user is called back}})
    {\If(\tcp*[f]{\texttt{\scriptsize if an explanation for $\tilde{y}_i$ is desired}}){$\mathit{is\_expl\_desired(x_i, \tilde{y}_i)}$}{
                $e_i \gets \mathit{get\_and\_show\_expl}(x_i, \tilde{y}_i, f, X_p)$;\tcp*[f]{\texttt{\scriptsize return explanation $e_i$}}\\
            }$\hat{y}_i \gets \mathit{h(x_i)}$; $y_i \gets \hat{y}_i$;   
    \tcp*[f]{\texttt{\scriptsize $\hat{y}_i$ accepted}}\\
    $h, f, p \gets \mathit{update}$;
    \tcp*[f]{\texttt{\scriptsize model and parameter updates}}\\
\lIf(\tcp*[f]{\texttt{\scriptsize phase change}}){$\mathit{check}(h,f, p)$}{
                    $phase = HiC, k=0$}

            }
            }

    ${X_p}, {Y_p} \gets \mathit{update (x_i, y_i)}$;
    \tcp*[f]{\texttt{\scriptsize recording final decision}}\\
                
\end{algorithm}

\section{A Bridget System}
\label{sec:method}
\bridget{}, whose pseudocode is reported in Algorithm~\ref{alg:algoFrank}, assumes two potential statuses, depicted in Figure~\ref{fig:bridget}: \begin{itemize}
    \item A %starting 
    \textit{Human-in-Command} (HiC) phase where %, as defined in~\cite{Mazzoni2024frank}, 
    the human $H$ and the machine $M$ are into a \textit{co-evolutionary relationship}~\cite{Mazzoni2024frank}. 
    $H$ takes all the decisions, and $M$ offering suggestions and explanations if skeptical. %, while increasingly becoming more aware of $H$'s behaviour. 
    \item A \textit{Machine-in-Command} (MiC) phase %, %triggered if certain conditions are met. 
    where $M$ takes most of the decisions, but it can call $H$ back if uncertain, and it is able to explain why.
\end{itemize} 

The \bridget{} system can loop between the two phases, accommodating the user's needs, potential fallouts in the model's accuracy, or novelties in the data. 

%RIC RIVEDERE DA QUI e due
\subsection{Human-in-Command} 
%At the beginning, \bridget{} is set in its HiC 
\bridget{} starts in the HiC phase and requires a set of records $X$, to be label one by
%which are received one by one, 
%and a transition condition $\mathit{t_M}$, leading to the MiC phase. %other phase. 
Once a new record $x_i$ is received, $H$ makes its decision as well as $M$, following SL (line 4 of Alg.~\ref{alg:algoFrank}). 
%Our implementation divergences in
%At this stage, we are very general about the stopping condition $\mathit{stp}$ as it might be implemented as reaching a certain number of labeled records, or an accuracy higher than a threshold\footnote{\scriptsize In our experiments, we consider as $\mathit{stp}$ a certain number of instances to be analyzed, leaving for future work the study of measures automatically unveiling when to stop the training.} for $f$.
%The system follows SL in this early HiC phase, with one key variation w.r.t. 
%how skepticality is computed. 
To compute skepticality, we propose \textit{Fading Empirical Accuracy} (FEA) as a replacement for EA. 
While computing $M$ and $H$'s track record, instead of assigning the same weight to each previously-seen record and the respectively assigned label, our metric weights each record w.r.t. its temporal distance to the current one (remember that $\tilde{Y}$, $\hat{Y}$, $Y$ and the other sets are ordered w.r.t. their appearance). 
In other words, older records weigh less. 
This is consistent with the idea of EA as an ever-evolving, more accurate proxy for the model's epistemic uncertainty, i.e., its \textit{current} status. 
For example, with FEA the model's early errors are de-emphasized. 
FEA is employed for $H$ as well, as their behavior might also change over time. 
Then, we define the \textit{Fading Skepticality} of $M$ towards the user decision $\hat{y}_i$ as:
%\begin{equation}
%    \label{eq:f_skept}
% $$
%     \mathit{fskp}(x_i, \tilde{y}_i, \hat{y}_i, Y, \tilde{Y}, \hat{Y}) = \mathit{c}(f, x_i, \tilde{y}_i) \cdot \mathit{fea}(\tilde{y}_i, Y, \tilde{Y}) - \mathit{c}(f, x_i, \hat{y}_i) \cdot \mathit{fea}(\hat{y}_i, Y, \hat{Y})
% $$
$$
    \mathit{skp}(x_i, \tilde{y}_i, \hat{y}_i,X_p, Y_p) = \mathit{c}_f(x_i, \tilde{y}_i) \cdot \mathit{fea}_f(x_i, \tilde{y_i}, X_p, Y_p) - \mathit{c}_f(x_i, \hat{y}_i) \cdot \mathit{fea}_h(x_i, \hat{y_i}, X_p, Y_p)
$$
%\end{equation}
\noindent where $\mathit{fea}_f(x_i, \tilde{y_i}, X_p, Y_p) = \frac{1}{|X_p|} \sum_{\{x_p \in X_p | f(x_p) =\tilde{y}_i\}}\delta_{f(x_p),y_p}d(x_p,x_i)$ and by analogy $\mathit{fea}_h(x_i, \hat{y_i}, X_p, Y_p) = \frac{1}{|X_p|} \sum_{\{x_p \in X_p | h(x_p) =\hat{y}_i\}}\delta_{h(x_p),y_p}d(x_p,x_i)$. Here $d(x_p,x_i)$ is a distance function between the current data point $x_i$ and previously seen data point $x_p$.
If $\mathit{fskp}$ is higher than a certain threshold $\alpha$, $M$ is skeptical of $H$'s decision (line 5), and the user is prompted if they want to change them.

Before making the final decision, $H$ can request an explanation for the suggestion, such as (counter-)exemplar records, local decision rules or an overview of the model's logic (lines 6-7). 
As in~\cite{Mazzoni2024frank}, the explanations are generated following the latest updated version of the model and the training data, and thus evolving over time. 
By exploiting in \bridget{} the \textsc{CAIPI} model~\cite{teso2019explanatory}, $H$ can also teach $M$ if the decision is right but for the wrong reason.
After a decision is taken, $M$'s internal model $f$ and the various sets are updated (line 11), following $H$'s decision, who always holds full veto power. 
This co-evolutionary phase ensures a profitable human-machine interaction both for $H$, as they might receive useful suggestions, and $M$, since the model is progressively updated.

As mentioned, for traditional EA, both $M$ and $H$ have as many FEA values as $|L|$, i.e., $2$ in a binary classification task. 
As SL assumes an expert user, the average of the model's FEA values can provide an esteem of its overall closeness to the user at any given point in time. 
As such, FEA also plays a key role in transitioning towards the machine-led phase, and after each labelled record, \bridget{} checks if a transition towards the MiC phase is possible (line 12). 
The transition happens if all the following set of conditions $\mathit{t_M}$ is met:
\begin{enumerate}
    \item during the co-evolutionary phase, more than $k_{max}$ records have been seen; 
    \item the model's average FEA is higher than a certain threshold;
    \item $H$ designates $M$ as the primary decision-maker.
\end{enumerate}
The first point employs SL's co-evolutionary phase as the traditional ML training step. 
Compared to LtD training, it assumes only one user, i.e., $H$, who is deemed reliable and who provides knowledge to $M$. 
The second point focuses on the quality of the model itself, i.e., to what extent it is aligned with the final decisions taken by $H$ at the current time. 
Combined, those two points lead to avoiding training a separate deferral system for the MiC phase, as it is assumed $M$ is a good approximation of $H$. 
With that said, the third point ensures $H$ willingly agrees with putting the machine in command. 

%, in order to later employ the model in \bridget{}'s MiC phase. 
%In the context of HDM, it can be seen as a counterpart of LtD training, albeit without the assumption of a potentially unreliable user with their own ground truth in the training of a deferring mechanism. 
%From the user's point of view, we believe this makes the model's training clearer, removing the multiple ground truths and the involvement of another, potentially unknown, third-party agent, focusing the process on the user-machine relationship instead. 
%The second point ensures the quality of the current state of the model. 

\subsection{Machine-in-Command} 
If $\mathit{t_M}$ is triggered, \bridget{} transitions to a state where $M$ is in command, labelling incoming records individually. 
%The system has a new transition condition $\mathit{t_H}$ to return to its original human-led state. 
As soon as a new record $x_i$ is received, $M$'s $f$ computes its \textit{belief}, i.e., its prediction probability, $b \in [0,1]$ towards its prediction (line 14), following the first part of the fading skepticality:
%\begin{equation}
%    \label{eq:b}
$$
    \mathit{b}(x_i, \tilde{y}_i, Y, \tilde{Y}) = \mathit{c}_f(x_i, \tilde{y}_i) \cdot \mathit{fea}_f(x_i, \tilde{y_i}, X_p, Y_p)
$$
%\end{equation} 

If $b$ is lower than a user-defined threshold $\beta$, $M$'s stance towards $x_i$ is considered unreliable, and $H$ is immediately called back to make their own decision $\hat{y}_i$ (lines 15 and 17). 
In this state, $M$ can explain why it is unreliable by providing, for instance, a small set of real or synthetic records with a low prediction probability similar to that achieved on $x_i$ (line 16).

Otherwise, if $b > \beta$, in order to prevent machine overreliance~\cite{rastogi2022deciding}, i.e., a common drawback of MiC systems, \bridget{} engages the user to check the model prediction on randomly selected record, not only those with a low belief.
As a simple implementation, \bridget{} draws a random number $r \in [0, 1]$.
If $r < \beta$, $M$'s prediction $\tilde{y}$ is accepted as $x_i$'s label (line 18), otherwhise the user $H$ is called back (line 19). Also, in the MiC phase, due to the reasonable level of reliability according to the high $b$, \bridget{} can provide the user with the same forms of explanations as in the HiC phase (lines 20-21).

Finally, the behaviour of $M$'s model $f$ after accepting a label differs between the HiC and MiC phases. 
Whereas $f$ is always updated after each decision originated either from $H$ or $M$ in the HiC phase, in the MiC phase
%the model tends to stay static in the MiC phase, as most LtD implementations.
$f$ and the various sets are only updated
%in line 19, after calling the user back. 
when the user is called back (line 23). 
This allows \bridget{} to notify the user if it reaches a critical status, i.e., if at least one of the following conditions is met (line 24):
\begin{enumerate}
    \item the user has been called back due to $M$'s low $b$ for more than $p_{max}$ times;
    \item the model's average FEA is now lower than a certain threshold.
\end{enumerate}
The user can then decide to come back in command, reverting to the HiC phase. 
These checks indicate a decrease in model reliability, potentially implying concept-drift.
%Those points show that the model is no longer reliable for current records or aligned with the user, potentially implying that a concept drift has occurred since the co-evolutionary phase, and newer records are near the decision border or in unexplored areas. 
Effectively, this doubles as a function commonly found in LtD systems, i.e., rejection of novelties and ambiguities~\cite{punzi2024ai}. 

\section{Conclusion and Future Works}
\label{sec:conclusion}

\begin{comment}

\begin{table}[t]
\centering
\caption{Differences between the phases.}
\begin{tabular}[t]{lcc}
\toprule
&\textbf{Human-in-Command}& \textbf{Machine-in-Command}\\
\midrule
Decisions& Most by the $H$, always vetting& Most by the $M$ and automated\\
Interaction&If model is skeptical&If model doubts of itself\\
Model Update&Always&When $H$ is called\\
%Explanations&Counter/factuals, Rules, Model logic&Guidance (Hints)\\
\bottomrule
\label{tab:com}
\end{tabular}
\end{table}%

\end{comment}

%Tab. \ref{tab:com} provides a comparison of \bridget{}'s two potential statuses.
% We have presented \bridget{}, our attempt to bridge the two major Hybrid Decision-Making paradigms. 
% \bridget{} leverages the co-evolutionary process commonly seen in SL to train a machine learning model as close as possible to the user's behaviour, which is then employed independently from the user themselves. 
% The model can also offer suggestions to the user or explain why it is in an unreliable state.
% The interaction between the two agents leads to a constant update of the system's parameters and, thus, a switch between the human-led and machine-led phases. 
% \bridget{} fully absorbs the Skeptical Learning paradigm, and compared to traditional Learning-to-Defer implementations, it removes the need for multiple ground truths and mitigates the user's overreliance on the machine.

We have presented \bridget{}, an approach designed to bridge the two main Hybrid Decision-Making paradigms. 
\bridget{} uses a co-evolutionary process %typical in SL 
to train a ML model to closely mimic the user's behavior. 
%Once trained, the model can operate independently of the user but also has the capability to provide suggestions or explain its limitations. 
This interactive dynamic between the human and machine agents allows for continual system parameter updates, resulting in alternating phases where either the human or the machine takes the lead. 
%\bridget{} fully embraces SL and improves upon traditional LtD by eliminating the need for multiple ground truths and reducing the user's potential overdependence on the machine.
%
%In the future, 
We plan to extensively test \bridget{} against stand-alone LtD systems and also consider in the implementation different data types, such as time series easily providing alternative types of explainability~\cite{spinnato2024understanding}, and more in-depth functions such as fairness checks~\cite{Mazzoni2024frank}. 
Moreover, we reckon \bridget{} should give a deeper focus on concept drift during phase transitions, as shifts in data are common reasons for a model's fall down.
Lastly, the current iteration of \bridget{} was designed around two principles -- employing FEA values as a model-agnostic proxy of the model's current status, and avoiding training an independent deferral system. 
Other approaches are possible, e.g., supplanting FEA values with the model's internal epistemic uncertainty or comparing the number of leaves of an incremental decision tree at two different points in time to assess the model's changes. 
Moreover, the user-provided decisions could effectively be used to train a deferral system at the end of the co-evolutionary phase. 
%attempting to predict when the machine would not agree with the human in the MiC phase.

{
\scriptsize
\textbf{Acknowledgment.}
This work is partially supported by the EU NextGenerationEU programme under the funding schemes PNRR-PE-AI FAIR, PNRR-SoBigData.it - Prot. IR0000013, H2020-INFRAIA-2019-1: Res. Infr. G.A. 871042 \textit{SoBigData++}, TANGO G.A. 101120763, and ERC-2018-ADG G.A. 834756 \textit{XAI}.
}

\bibliographystyle{abbrv}
\bibliography{main}

\end{document}